\documentclass[letterpaper, 10 pt, conference]{ieeeconf}

\usepackage{censor}
\usepackage{cite}
\usepackage{graphicx}
\usepackage{amsmath, amssymb}
\usepackage{algorithm}
\usepackage{algpseudocode}
\usepackage{comment}
\usepackage{xcolor}
\usepackage{booktabs}
\usepackage{dsfont}
\usepackage[hyphens]{url}
\usepackage{xurl}
\usepackage[unicode, hidelinks, breaklinks=true]{hyperref}

\IEEEoverridecommandlockouts
\title{\LARGE \bf
HapticVLA: Contact-Rich Manipulation via Vision-Language-Action Model without Inference-Time Tactile Sensing
}

\author{
    \parbox{18cm}{\centering
        Konstantin Gubernatorov*, Mikhail Sannikov*, Ilya Mikhalchuk*, Egor Kuznetsov, Makar Artemov, \\
        Ogunwoye Faith Oluwatobi, Marcelino Julio Fernando, Artem Asanov, Ziang Guo, Dzmitry Tsetserukou
    }
    \thanks{*Denotes equal contribution.}
    \thanks{All authors are with the Intelligent Space Robotics Laboratory, Skolkovo Institute of Science and Technology, Moscow, Russia. \tt \{{Konstantin.Gubernatorov}, {Mikhail.Sannikov}, {Ilia.Mikhalchuk}, {Egor Kuznetsov}, {Makar.Artemov}, {Faith.Ogunwoye}, {Marcelino.Fernando}, {Artem.Asanov}, {Ziang.Guo}, {D.Tsetserukou}\} @skoltech.ru}
}

\begin{document}

\maketitle
\thispagestyle{empty}
\pagestyle{empty}

\begin{abstract}

Tactile sensing is a crucial capability for Vision-Language-Action (VLA) architectures, as it enables dexterous and safe manipulation in contact-rich tasks. However, reliance on dedicated tactile hardware increases costs and reduces reproducibility across robotic platforms. We argue that tactile-aware manipulation can be learned offline and deployed without direct haptic feedback at inference. To this end, we present HapticVLA, which proceeds in two tightly coupled stages: Safety-Aware Reward-Weighted Flow Matching (SA-RWFM) and Tactile Distillation (TD). SA-RWFM trains a flow-matching action expert that incorporates precomputed, safety-aware tactile rewards, penalizing excessive grasping force and suboptimal grasping trajectories. TD further transfers this tactile-aware capability into a conventional VLA: we distill a compact tactile token from the SA-RWFM teacher and train a student VLA to predict that token from vision and state modalities, enabling tactile-aware action generation at inference without requiring on-board tactile sensors. This design preserves contact-rich tactile-aware reasoning within VLA while removing the need for on-board tactile sensors during deployment. In real-world experiments, HapticVLA achieves a mean success rate of 86.7\%, consistently outperforming baseline VLAs — including versions provided with direct tactile feedback at inference. All code, models, datasets, and a digital twin of our setup are available on the \href{https://advanced-robotic-manipulation.github.io/websites/websites/hapticvla/}{\textcolor{blue}{Project website}}.

\end{abstract}

\section{Introduction}

\begin{figure}[t]
    \centering
    \includegraphics[width=\columnwidth]{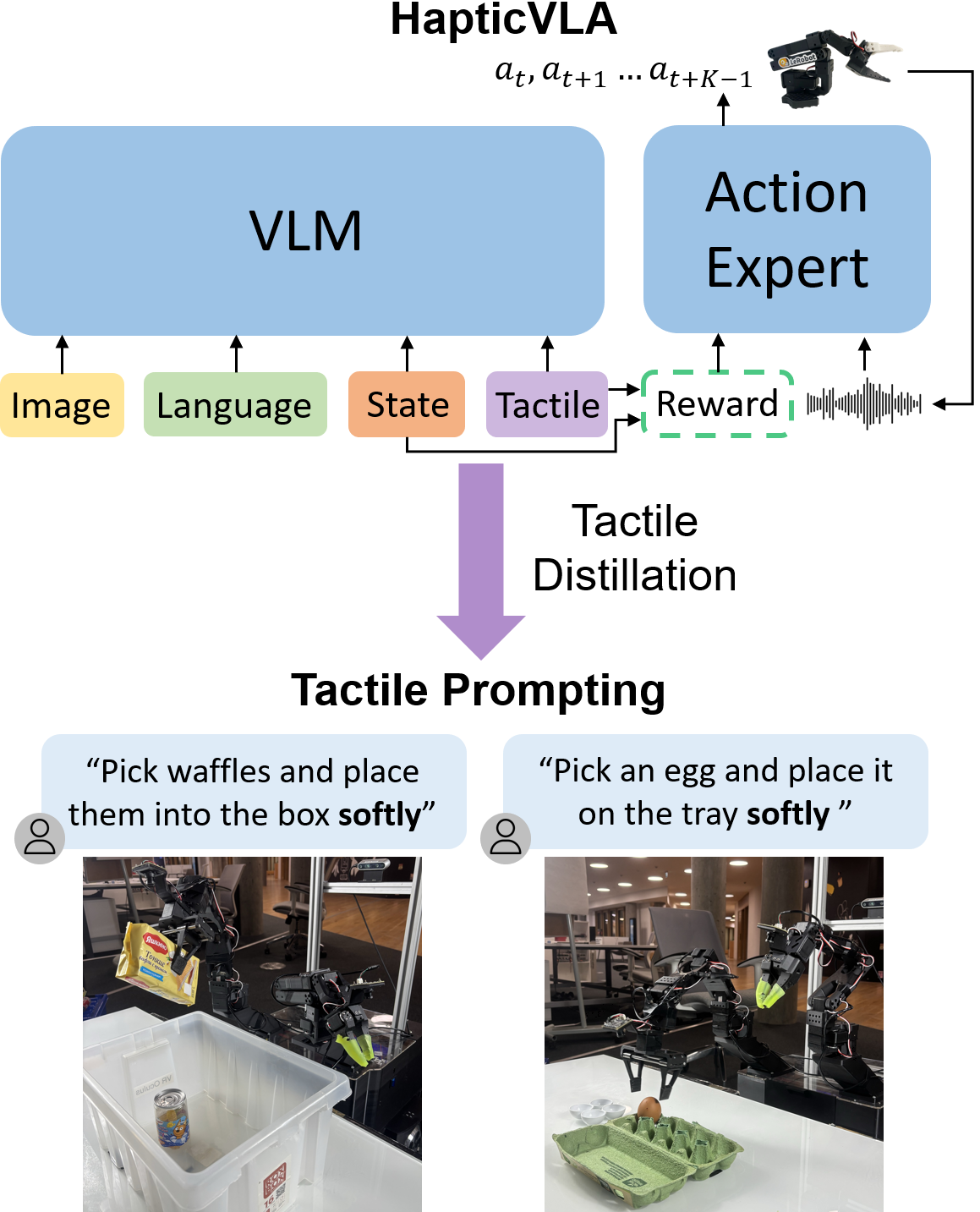}
    \caption{HapticVLA is the first VLA model to enable tactile-aware manipulation without requiring force sensors at inference. Alongside vision, language, and state modalities it processes tactile reward, enabling safe action generation for contact-rich manipulation. Our approach leverages tactile distillation to preserve haptic-aware reasoning while eliminating tactile modality through distillation.}
    \label{fig:teaser}
\end{figure}

Vision-Language-Action models (VLAs) unify visual perception and language understanding to generate robot control commands that enable scalable manipulation of everyday tasks, including complex actions that require bimanual or even mobile bimanual setups~\cite{kimopenvla, black2024pi0visionlanguageactionflowmodel, shukor2025smolvla}. They take visual input from cameras and proprioceptive states, and output continuous motor commands. Hence, they rely solely onvision for environment perception, lacking explicit force or tactile awareness. In practice, this means that a VLA cannot understand hardness or solidity, severely limiting safe, contact-rich manipulation of soft and fragile objects.

This limitation motivated recent VLA models to incorporate tactile or force modalities, aligning them with language and vision modalities as input into Vision-Language Model (VLM). The majority of approaches enable this by implementing visual-force sensors on the fingers of grippers~\cite{zhang2026vtla, liu2025mla, cheng2026omnivtla, bi2026vla, 10.1145/3728485.3759237, jones2025beyond}. These sensors use an internal camera to capture surface deformations when the robot makes contact by measuring contact forces, textures, and material properties that vision alone cannot infer. However, the proposed methods encode raw sensor signals as a vision modality, overlooking the unique modality of tactile sensing: unlike scene vision, which captures remote photometric properties, tactile sensing captures local mechanical interactions. Hence, treating tactile data solely as a vision modality fails to extract the contact force information critical for tactile-aware manipulation tasks.

Another way to perform contact-rich manipulation is to explicitly learn force from  force sensors in the motors in the joints of the manipulators~\cite{yu2026forcevla} and combine both tactile and force modalities~\cite{huang2026tactile}. This enables VLAs not only to grasp objects with a wide range of material properties but also to perform tasks requiring varying pressure force, such as insertion or wiping surfaces of different smoothness. FD-VLA also follows the force-based approach, but similar to one of our contributions, eliminates the need for force sensors during inference while still performing force-aware manipulation~\cite{zhao2026fd}. The proposed distillation of force data into a learnable token, aligning it with vision and state modalities, shows promising accuracy for contact-rich manipulation without any force/tactile sensors.

\begin{figure*}[t]
    \centering
    \includegraphics[width=\textwidth]{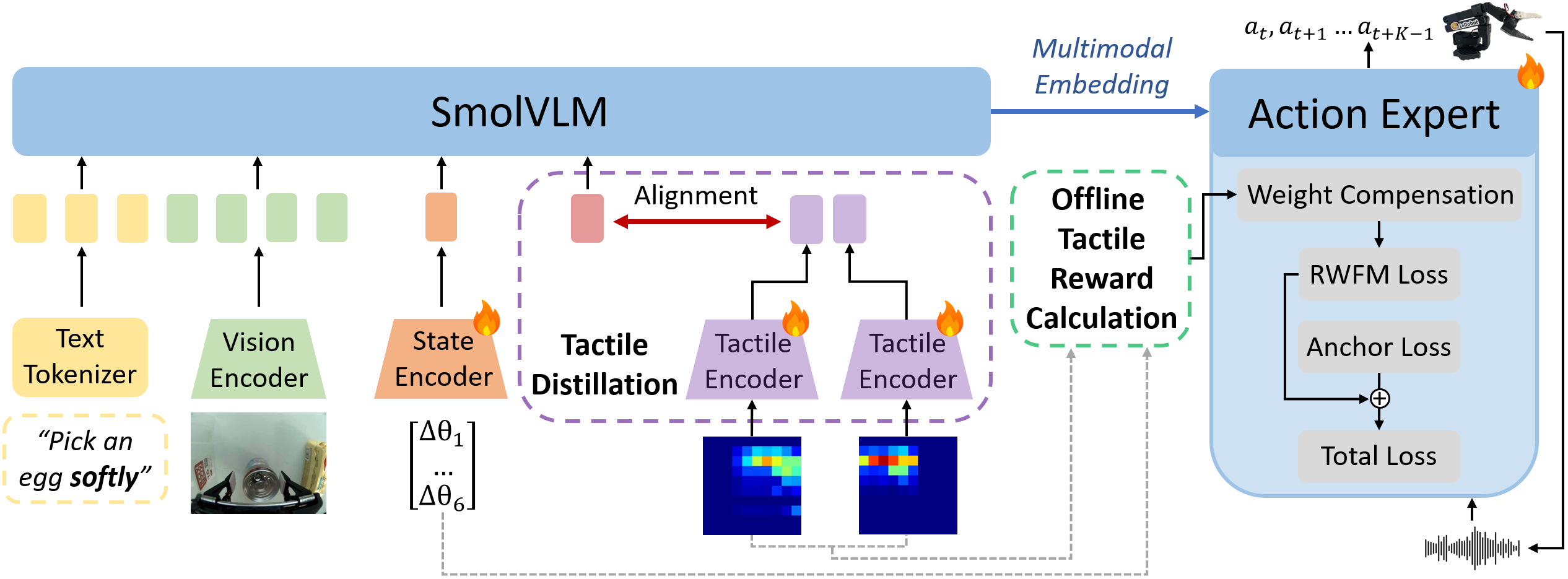}
    \caption{Framework overview of HapticVLA. After dataset collection we perform offline tactile reward calculation for each episode based on manipulator state and tactile maps. During VLA training, along with language, vision and state modalities passed into VLM, this reward is passed as an additional condition directly into SA-RWFM action expert. To eliminate the need for tactile hardware, we further perform tactile distillation on our trained VLA. At inference, the distilled VLA predicts tactile representations solely from vision and state inputs enabling tactile-aware contact-rich manipulation.}
    \label{fig:method}
\end{figure*}

In this work, we propose a novel HapticVLA framework that incorporates a Safety-Aware Reward-Weighted Flow Matching (SA-RWFM) action expert for encoding tactile-aware safe contact-rich manipulation abilities into VLA and a Tactile Distillation (TD) framework that incorporates a distilled tactile token, rather than raw tactile sensor feedback, into the VLA model. As shown in Fig.~\ref{fig:method}, our proposed approach comprises three main stages: (1) \textit{Offline Tactile Reward Calculation}, where we compute rewards for each collected episode based on the state of the manipulator and tactile feedback, penalizing high grasping forces applied to fragile objects; (2) \textit{SmolVLA} with SA-RWFM action expert training ~\cite{shukor2025smolvla}, where we fine-tune VLA incorporating tactile sensors feedback and aligning it with state and vision modalities; (3) \textit{TD framework}, in which the SA-RWFM-trained SmolVLA is distilled into a conventional SmolVLA enabling tactile-aware reasoning without requiring direct sensor measurements. This approach enables real-world deployment on a wide range of robotic setups that lack tactile sensors, reducing hardware complexity and cost while still benefiting from tactile-aware VLA manipulation.

\section{Related Work}

\subsection{VLAs for Contact-Rich Manipulation}

Recent works enhance conventional VLA models by integrating tactile sensing, improving contact-rich manipulation. This is achieved by leveraging visual-force sensors on the fingers of parallel grippers encoding their outputs similarly to the vision modality from cameras. Tactile-VLA~\cite{huang2025tactile} fuses vision, language, action, and tactile feedback, using a hybrid position–force controller and a reasoning module to adapt to tactile cues. Similarly, VTLA~\cite{zhang2026vtla} jointly processes visual and tactile inputs with language grounding, using a Direct Preference Optimization (DPO) loss for insertion tasks. OmniVTLA~\cite{cheng2026omnivtla} employs a dual-path tactile encoder and a large text–vision–tactile dataset to train a semantically-aligned tactile representation, yielding much higher success rates on pick-and-place tasks. VLA-Touch~\cite{bi2026vla} adds a pretrained tactile–language model for semantic feedback alongside a diffusion-based controller that refines VLA actions with real tactile signals. Liu et al. introduced MLA~\cite{liu2025mla}, a multisensory VLA that integrates 2\,D images, 3\,D point clouds, and tactile signals through an encoder-free alignment mechanism that repurposes the LLM itself for perception. By forecasting future multisensory states, the model reasons about physical dynamics to regulate interaction forces in force-sensitive tasks like whiteboard wiping and tool-based assembly, outperforming visual-only methods in real-world interactions. However, all of these approaches require high-end visual-tactile sensors, severely limiting reproducibility due to the high cost and compatibility issues with commonly used grippers and manipulators.

Other works explicitly incorporate force feedback from motors in manipulator joints. ForceVLA~\cite{yu2026forcevla} treats external force as a first-class modality via a Mixture-of-Experts fusion. It augments vision-language embeddings with 6-axis force inputs during action decoding, achieving an improved success rate in insertion tasks. TaF-VLA~\cite{huang2026tactile} proposes a Tactile-Force Adapter trained on a 10M-sample dataset, aligning tactile inputs to actual force measurements rather than vision modality, producing force-aware manipulation in tasks with varying grasping force and different hardness of objects and surfaces.

\subsection{Contact-rich Manipulation without Force Sensors at Inference}
Gano et al. show that pre-training a vision policy with low-fidelity tactile sensors and then disabling them at inference dramatically improves vision-only manipulation for insertion tasks~\cite{gano2025low}. Beyond Sight~\cite{jones2025beyond} fine-tune generalist visuomotor policies on visual, vision-based tactile, and audio modalities using language as a cross-modal grounding signal. These methods demonstrate the benefit of adding tactile data during training for tasks that involve objects of different hardness and insertion. FD-VLA~\cite{zhao2026fd} learns force awareness without force sensors at inference: it uses a Force Distillation Module, which distills force by mapping a learnable query token, conditioned on visual observations and robot states, to a predicted force token aligned with the latent representation of actual force signals. During inference, this distilled force token is injected into the pretrained VLM, enabling force-aware reasoning without any force feedback via the alignment with visual-language semantics. The authors found that the learnable distilled force token outperforms approaches using tactile or force feedback during inference, as well as conventional VLA architectures.

\subsection{Reinforcement Learning for Flow Matching VLA Policies}

Flow matching (FM) policies are now central to VLAs, and several works introduce Reinforcement Learning (RL) based fine-tuning to enhance these policies. Pfrommer et al. propose two methods to improve flow policies beyond imitation: Reward-Weighted Flow Matching (RWFM) and Group Relative Policy Optimization (GRPO)~\cite{pfrommer2025reinforcement}. The authors use RL signals to exceed suboptimal demonstrations, incorporating variable-horizon planning into the flow model. Fan et al. describe an online fine-tuning method that integrates reward-weighting into FM while adding a Wasserstein regularization to preserve diversity~\cite{liu2026flow}. Their approach guides the flow model toward high-reward regions without requiring reward gradients, balancing exploitation with a computationally tractable diversity penalty. Zhu et al. propose a different approach in FlowRL~\cite{zhu2025flowrl}: instead of maximizing the reward, they match the entire reward distribution through a flow-based objective.

Specifically in robotics applications, Zhang et al. propose Adaptive Reinforced Flow Matching (ARFM) as an offline RL post-training for VLA flows~\cite{zhang2026balancing}. ARFM adds an adaptive scaling factor to the loss, removing bias and variance to stabilize fine-tuning. The authors achieve better accuracy and robustness on downstream tasks. FlowRAM~\cite{wang2025flowram} focuses on perception efficiency by leveraging region-based generative modeling and learning actions through conditional FM. It achieves SOTA performance in high-precision tasks with very low planning time. Zhang et al. introduce ReinFlow~\cite{zhang2026reinflow}, an online RL framework that injects noise into a flow policy to form a discrete Markov process. ReinFlow fine-tunes various flow variants, showing dramatic gains (+135\% reward in locomotion, +40\% success in manipulation) against diffusion PPO baselines.

Following the success of offline RL for FM VLAs, our work leverages this approach to incorporate tactile data and contact-rich safety rewards into FM objectives.

\subsection{Cross-modal Knowledge Distillation for Efficient VLAs}

Knowledge distillation (KD) has emerged as a primary mechanism for transferring the reasoning capabilities of high-capacity VLAs into compact policies suitable for real-time deployment on edge devices. Recent research focuses on architectural and task-aware optimizations: X-Distill~\cite{shao2026xdistill} enables cross-architecture transfer from large Vision Transformer~\cite{dosovitskiyimage} teachers to ResNet-18 students to leverage local inductive biases, demonstrated via simulated tasks in MetaWorld~\cite{yu2019metaworld}. ActDistill~\cite{ye2025actdistill} further optimizes inference through action-guided self-derived distillation, utilizing graph-structured encapsulation and dynamic routers to reduce computation by more than 50\% on the LIBERO~\cite{liu2023libero} and CALVIN~\cite{mees2022calvin} benchmarks. Complementing these structural approaches, DySL-VLA~\cite{yang2026dyslvla} introduces skip-aware distillation for dynamic layer skipping based on action importance, achieving a 3.75$\times$ speedup in CALVIN and deployment on Jetson Orin. For multi-modal systems, MoVE-KD~\cite{cao2025movekd} distills the proficiencies of diverse visual encoders into a single student using a Mixture-of-Experts strategy integrated into the RoboBrain model~\cite{ji2025robobrain} for complex planning, affordance perception, and trajectory prediction. FedVLA~\cite{miao2025fedvla} extends this to decentralized settings by facilitating cross-client knowledge transfer through expert-driven aggregation in a federated learning framework for privacy-preserving robotic manipulation.

\section{Method}

\newcommand{\R}{\mathbb{R}}
\newcommand{\I}{\mathbb{1}}
\newcommand{\relu}{\mathrm{ReLU}}
\newcommand{\clip}{\mathrm{clip}}
\newcommand{\eps}{\varepsilon}
\newcommand{\argmin}{\operatorname*{arg\,min}}
\newcommand{\argmax}{\operatorname*{arg\,max}}
\newcommand{\E}{\mathbb{E}}
\newcommand{\Var}{\mathrm{Var}}
\newcommand{\norm}[1]{\left\lVert #1\right\rVert}
\newcommand{\mad}{\mathrm{MAD}}

\subsection{Safety-Aware Reward-Weighted Flow Matching}

\subsubsection{Safety Reward from Tactile Maps}

We compute a tactile-based safety reward that is used only during offline fine-tuning. At each step $t$, from tactile sensors left and right tactile maps $\mathbf{M}_t^L,\mathbf{M}_t^R\in[0,1]^{H\times W}$ are received, where $H$ and $W$ are the dimensions of the tactile maps.

\paragraph{Normalization, activation, and calibration}
Raw tactile readings are normalized using a dataset-level scale estimated from the 99th percentile of the tactile signal.
To ensure robustness across all episodes in the dataset, all reward thresholds are calibrated from contact statistics in the dataset with quantiles: a safe force band $[f_{min},f_{max}]$, a peak-pressure limit $p_{max}$, a concentration limit $c_{max}$, and a tolerance of inter-pad force asymmetry $\delta$ when both pads are active.
We denote by $\mathcal{A}\subseteq\{L,R\}$ the set of active tactile sides after this calibration.

\paragraph{Contact statistics}
For each active side $s\in\mathcal{A}$, we summarize the tactile map by mean force proxy $f_t^s$, peak pressure $p_t^s$, and pressure concentration $c_t^s$:

\begin{align}
f_t^s=\frac{1}{HW}\sum_{i,j} M_{t,ij}^s, \\
p_t^s=\max_{i,j} M_{t,ij}^s, \\
c_t^s=\frac{p_t^s}{HW\cdot f_t^s+\eps}.
\end{align}

We also compute the center of pressure (CoP) on a normalized grid $(X_{ij},Y_{ij})\in[0,1]^2$:
\begin{equation}
\mathrm{cop}_{x,t}^s=\frac{\sum_{i,j}M_{t,ij}^sX_{ij}}{\sum_{i,j}M_{t,ij}^s+\eps},\quad
\mathrm{cop}_{y,t}^s=\frac{\sum_{i,j}M_{t,ij}^sY_{ij}}{\sum_{i,j}M_{t,ij}^s+\eps}.
\end{equation}

\paragraph{Holding and slip}
We define a binary holding state $h_t$ as stable contact while the gripper remains closed.
Slip is detected only during holding, based on either a CoP jump or a sudden force drop:
\begin{equation}
\mathrm{slip}_t= \mathds{I} [h_t=1]\cdot \mathds{I} \Big[
\max_{s\in\mathcal{A}}\Delta \mathrm{cop}_t^s>c_{op}\ \lor\
\min_{s\in\mathcal{A}}\Delta f_t^s<-d_f
\Big],
\end{equation}
where
\begin{equation}
\Delta \mathrm{cop}_t^s=
\left\lVert
(\mathrm{cop}_{x,t}^s,\mathrm{cop}_{y,t}^s)-
(\mathrm{cop}_{x,t-1}^s,\mathrm{cop}_{y,t-1}^s)
\right\rVert_2
\end{equation}

\subsubsection{Per-step Reward and Episode Risk}
The per-step reward penalizes unsafe tactile patterns (over-force, under-force while holding, peak pressure, concentration, asymmetry, and slip):
\begin{equation}
\small
\begin{aligned}
r_t =&-\sum_{s\in\mathcal{A}}\Big(
\lambda_{high}\relu(f_t^s-f_{max})^2 \\
&+\lambda_{low} \mathds{I} [h_t=1]\relu(f_{min}-f_t^s)^2 \\
&+\lambda_{peak}\relu(p_t^s-p_{max})^2 \\
&+\lambda_{conc}\relu(c_t^s-c_{max})^2 \Big)\\
&-\lambda_{asym} \mathds{I} [L,R\in\mathcal{A}]\,\relu(|f_t^L-f_t^R|-\delta)^2 \\
&-\lambda_{slip}\,\mathrm{slip}_t.
\end{aligned}
\end{equation}

To capture near-failure behavior over the full trajectory, we also compute an episode-level risk score.
Let $e_t$ denote the maximum normalized threshold exceedance across
force, peak pressure, and concentration, and let $\mathcal{M} \subseteq
\{1, \dots, T\}$ be the set of timesteps $t$ with positive holding
state $h_t$, where $T$ is the full episode length. We define
\begin{equation}
\mathrm{risk}=\clip\!\left(P_{95}(\{e_t:t\in\mathcal{M}\})+\tfrac{1}{2T}\sum_{t=1}^{T}\mathrm{slip}_t,\ 0,\ 1\right).
\end{equation}

\subsubsection{Episode Reward}
The final episode reward combines the averaged step reward, task outcome, and safety risk penalty:
\begin{equation}
R_{step}=R_{step\_scale}\frac{1}{T}\sum_{t=1}^{T} r_t
\end{equation}
\begin{equation}
\begin{aligned}
R_{episode}
&= R_{step}
+ R_{succ}\,\texttt{success}
- R_{drop}\,\texttt{drop} \\
&\quad - R_{damage}\,\texttt{damage}
- R_{risk}\,\texttt{risk}.
\end{aligned}
\end{equation}

\subsubsection{SA-RWFM Fine-Tuning}
\label{subsec:rwfm}
Given an offline dataset $\mathcal{D}={(o_t, a_{t:t+K-1}, \tau, r_t, R_{\text{episode}})}$, we fine-tune a flow-matching VLA policy by weighting each training sample according to its computed reward. Here, $o_t$ denotes the observation at time step $t$, $a_{t:t+K-1}$ is the action sequence over a horizon of $K$ steps, $\tau$ is the timestep of the dataset sample, $r_t$ represents the per-step reward, and $R_{\text{episode}}$ is the episode reward.

\paragraph{Per-sample FM loss with task masking}
We compute a masked per-sample loss:
\begin{equation}
L_i=\frac{\sum_{h,j} m_{i,j}\ell_{i,h,j}}{H\sum_j m_{i,j}+\eps},
\end{equation}
where $m_{i,j}\in\{0,1\}$ is the task-dependent degree of freedom mask.

\paragraph{Mixed local/global returns}
For a chunk starting at time $t$, we compute discounted chunk return:
\begin{equation}
R^{chunk}_t=\sum_{k=0}^{H-1}\gamma^k r_{K+k},\qquad \gamma=0.99,
\end{equation}
and use the episode reward $R_{episode}$ for global outcome credit.

\paragraph{Robust group-wise normalization}
To avoid scale bias across datasets, we compute robust normalized scores within group $g$:
\begin{equation}
z=\frac{x-\mathrm{median}_g(x)}{\mathrm{scale}_g}
\end{equation}
where:
\begin{equation}
\mathrm{scale}_g=\max(1.4826\cdot \mathrm{MAD}_g(x),\epsilon).
\end{equation}
with a fallback to standard deviation if Mean Absolute Deviation ($\mad$) is degenerate.

\paragraph{Advantage-like score}
We combine normalized episode and chunk scores:
\begin{equation}
A_i=\clip\left(\beta z^{epi}_i+(1-\beta)z^{chunk}_i,\ -c_A,\ c_A\right),
\end{equation}
with $\beta=0.7$ and $c_A=6$.

\paragraph{Exponentiated, clipped, renormalized weights}
We define raw weights
\begin{equation}
w_i^{raw}=\exp(\alpha A_i),
\end{equation}
clip $w_i^{clip}=\clip(w_i^{raw},w_{min},w_{max})$, and renormalize within the batch:
\begin{equation}
w_i=\frac{w_i^{clip}}{\frac{1}{B}\sum_{b=1}^{B}w_b^{clip}}.
\end{equation}
Our hyperparameters are $\alpha=0.25$, $w_{min}=0.25$, and $w_{max}=4.0$.

\paragraph{RWFM objective}
The reward-weighted FM loss is
\begin{equation}
L_{\text{rwfm}}=\frac{\sum_{i=1}^{B} w_i L_i}{\sum_{i=1}^{B} w_i}.
\end{equation}

\subsubsection{Anchor Regularization for Stability}
Exponentiated weighting can cause drift and mode collapse in generative policies \cite{liu2026flow, zhang2026balancing}.
We add an anchor regularizer that constrains the parameters to remain near the initial imitation solution $\theta_0$:
\begin{equation}
L_{\text{anchor}}=\frac{1}{|\Theta|}\sum_{p\in\Theta}\norm{\theta_p-\theta_p^{0}}_2^2,
\end{equation}
and optimize
\begin{equation}
L_{\text{total}} = L_{\text{rwfm}} + \lambda_{\text{anchor}} L_{\text{anchor}}.
\end{equation}
We apply a warm-up schedule that gradually increases $\alpha$ and adjusts $\lambda_{\text{anchor}}$ to preserve baseline behavior early in training.

Algorithm~\ref{alg:sa_rwfm} outlines the fine-tuning procedure of the SA-RWFM action expert.

\begin{algorithm}[t]
\caption{SA-RWFM Offline Fine-Tuning Step}
\label{alg:sa_rwfm}
\footnotesize
\begin{algorithmic}[1]
\State \textbf{Input:} batch $\{(o_i,\tau_i,a_i,\{r_{i,t}\},R^{epi}_i,g_i)\}_{i=1}^{B}$ with group label $g_i$
\State \textbf{Input:} base parameters $\theta_0$ and current parameters $\theta$
\State \textbf{Input:} hyperparameters $(\alpha,\beta,\gamma,c_A,w_{min},w_{max},\lambda_{\text{anchor}})$
\State Compute per-element FM loss tensor $\ell_{i,h,j}$ using the flow policy $\pi_\theta(a|o_i,\tau_i)$
\State Compute masked per-sample loss $L_i$ (Eq.~(11))
\State Compute chunk return $R^{chunk}_i=\sum_{k=0}^{H-1}\gamma^k r_{i,t+k}$
\State Use episode reward $R^{epi}_i$
\State Robustly normalize within group $g_i$ to get $(z^{chunk}_i,z^{epi}_i)$ (Eq.~(13))
\State Compute $A_i=\clip(\beta z^{epi}_i+(1-\beta)z^{chunk}_i,-c_A,c_A)$ (Eq.~(15))
\State $w_i^{raw}\leftarrow \exp(\alpha A_i)$;\quad $w_i^{clip}\leftarrow \clip(w_i^{raw},w_{min},w_{max})$
\State $w_i\leftarrow w_i^{clip}/\mathrm{mean}(\{w_b^{clip}\}_{b=1}^{B})$
\State $L_{\text{rwfm}}\leftarrow \frac{\sum_i w_i L_i}{\sum_i w_i}$
\State $L_{\text{anchor}}\leftarrow \frac{1}{|\Theta|}\sum_{p\in\Theta}\norm{\theta_p-\theta_p^{0}}_2^2$
\State $L_{\text{total}}\leftarrow L_{\text{rwfm}}+\lambda_{\text{anchor}}L_{\text{anchor}}$
\State Update $\theta \leftarrow \theta - \eta \nabla_\theta L_{\text{total}}$
\end{algorithmic}
\end{algorithm}

\subsection{Tactile Distillation}
\label{sec:distillation}

While the SA-RWFM teacher achieves tactile-aware manipulation, it requires physical tactile sensors at inference time, constraining deployment to hardware-equipped platforms. To decouple tactile knowledge from tactile hardware, we propose an offline TD framework that distills the teacher's tactile-aware actions into a conventional VLA student with zero architectural modifications.

\subsubsection{Problem Formulation}

Let $\pi_T(\mathbf{o}, \mathbf{s}_T)$ denote the tactile-conditioned teacher policy, where $\mathbf{o} = (\mathbf{I}, \ell)$ consists of a visual observation $\mathbf{I} \in \mathbb{R}^{H \times W \times 3}$ and a language instruction $\ell$, and $\mathbf{s}_T = [\mathbf{q}; \mathbf{f}] \in \mathbb{R}^{134}$ concatenates the proprioceptive joint state $\mathbf{q} \in \mathbb{R}^{6}$ with a tactile embedding $\mathbf{f} \in \mathbb{R}^{128}$ produced by the dual tactile encoder. 

Both policies parameterize action chunks $\mathbf{a}_{1:H} \in \mathbb{R}^{H \times d_a}$ via conditional flow matching, where $K = 50$ is the action horizon and $d_a = 6$ is the per-arm joint dimension.

\subsubsection{Stage 1: Offline Teacher Target Generation}

Rather than requiring the teacher at training time, we pre-compute teacher action predictions on the full training set once, making the distillation process completely offline. For each training sample $i$ with observation $\mathbf{o}_i$ and full state $\mathbf{s}_{T,i}$, we run teacher inference to obtain a predicted action chunk:
\begin{equation}
    \hat{\mathbf{a}}_i^T = \pi_T(\mathbf{o}_i, \mathbf{s}_{T,i}) \in \mathbb{R}^{H \times d_a}.
    \label{eq:teacher_pred}
\end{equation}
This produces a dataset of soft targets $\{\hat{\mathbf{a}}_i^T\}_{i=1}^{N}$ with $N$ samples. Since the teacher was trained with SA-RWFM, these predictions implicitly encode both the tactile-conditioned force awareness and the reward-weighted quality bias: action chunks corresponding to successful demonstrations carry higher fidelity than those from suboptimal episodes.

\subsubsection{Stage 2: Teacher Backbone Initialization}

To transfer learned visual and language representations, we initialize the student from the teacher's weights with targeted adaptation. Let $\Theta_T$ denote the teacher parameters. We copy all parameters except the tactile encoder:
\begin{equation}
    \Theta_S^{(0)} = \big\{\theta \in \Theta_T \mid \theta \notin \Theta_{\text{tactile}}\big\}.
\end{equation}
The state augmentation projection requires special handling since the input dimension changes. The teacher's projection $\mathbf{W}_T \in \mathbb{R}^{32 \times 134}$ maps the concatenated proprio-tactile state to the SmolVLA latent space. We extract the first $d_a = 6$ columns corresponding to proprioception:
\begin{equation}
    \mathbf{W}_S = \mathbf{W}_T[:, :d_a] \in \mathbb{R}^{32 \times 6}, \quad \mathbf{b}_S = \mathbf{b}_T \in \mathbb{R}^{32},
\end{equation}
where $\mathbf{b}_T$ is the shared bias. This preserves the teacher's learned mapping from joint positions to the latent space while discarding the tactile columns, giving the student a strong initialization aligned with the teacher's representation space.

\subsubsection{Stage 3: Blended Target Training}

The student trains on blended action targets that interpolate between ground-truth demonstrations and teacher predictions:
\begin{equation}
    \tilde{\mathbf{a}}_i = (1 - \alpha_d)\,\mathbf{a}_i^{\text{GT}} + \alpha_d\,\hat{\mathbf{a}}_i^T,
    \label{eq:blended_target}
\end{equation}
where $\alpha_d \in [0, 1]$ is the blending coefficient and $\mathbf{a}_i^{\text{GT}}$ is the ground-truth action chunk from the demonstration. The distillation loss substitutes $\tilde{\mathbf{a}}_i$ as the flow matching target:
\begin{equation}
    \mathcal{L}_{\text{distill}}(\theta) = \mathbb{E}_{t, \mathbf{x}_0, \tilde{\mathbf{a}}}\!\Big[\big\| v_\theta(\mathbf{x}_t, t) - (\tilde{\mathbf{a}} - \mathbf{x}_0) \big\|^2\Big].
    \label{eq:distill_loss}
\end{equation}

We set $\alpha_d = 0.5$, which provides an equal blend that balances two complementary signals:

\begin{itemize}
    \item Ground-truth anchoring ($1 - \alpha_d$): prevents the student from amplifying teacher errors, keeping actions grounded in demonstrated behavior.
    \item Teacher shaping ($\alpha_d$): injects tactile-aware force modulation learned by the SA-RWFM teacher.
\end{itemize}

During validation, we set $\alpha_d = 0$ to obtain an unbiased estimate of reconstruction quality, enabling fair comparison with non-distilled baselines.

\subsection{Dataset Collection}

To support our proposed SA-RWFM objective, we deliberately collected both successful and failure-mode demonstrations that enable explicit reward weighting based on tactile feedback and post-grasp object integrity.

For each single-arm pick-and-place task, we collected 70 successful episodes and 20 faulty episodes per task, where faulty episodes are those in which excessive grasping force produced permanent structural damage or significant object deformation. For the bimanual manipulation task, we collected 100 successful and 30 faulty episodes. In total, our in-house real-world dataset comprises 310 episodes across the three tasks. Alongside positive rewards from successful episodes, faulty demonstrations are assigned negative rewards derived from measured contact forces and manipulator state. These negative examples are crucial for the SA-RWFM action expert: they constrain and sharpen the high-reward region of the target distribution, improving the model's ability to discriminate safe, high-reward trajectories from unsafe ones.

\begin{figure}[t]
    \centering
    \includegraphics[width=\columnwidth]{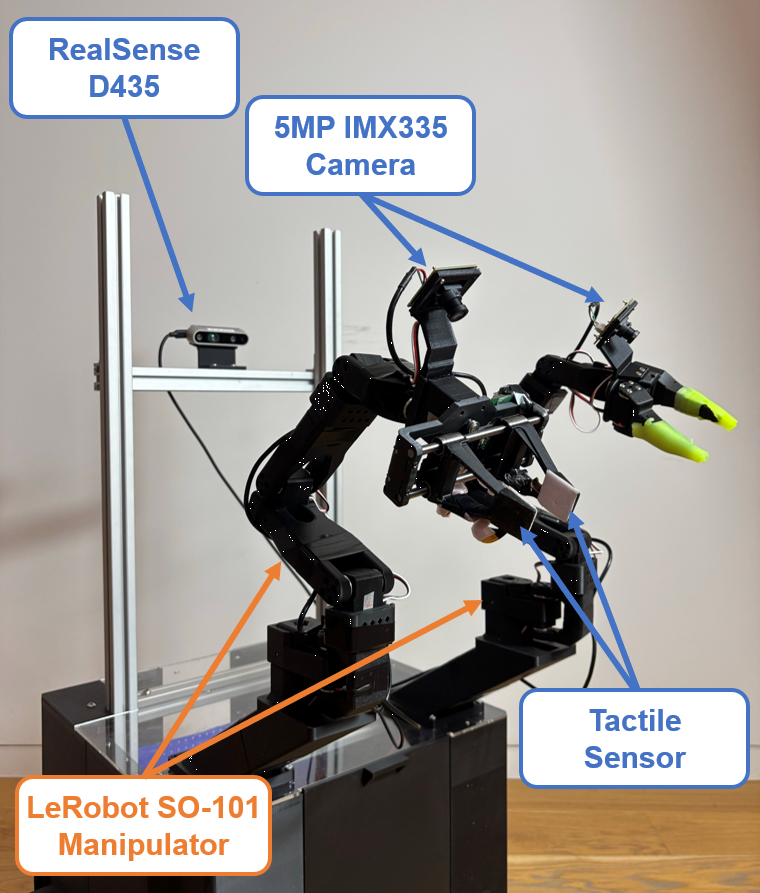}
    \caption{For our robotics platform we use two SO-101 robot arms as the main manipulation platform, an Intel RealSense D\,435 as the main camera and wrist-mounted IMX335 cameras.}
    \label{fig:robot}
\end{figure}

\begin{figure}[t]
    \centering
    \includegraphics[width=0.4725\columnwidth]{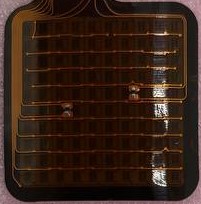}
    \hfill
    \includegraphics[width=0.48\columnwidth]{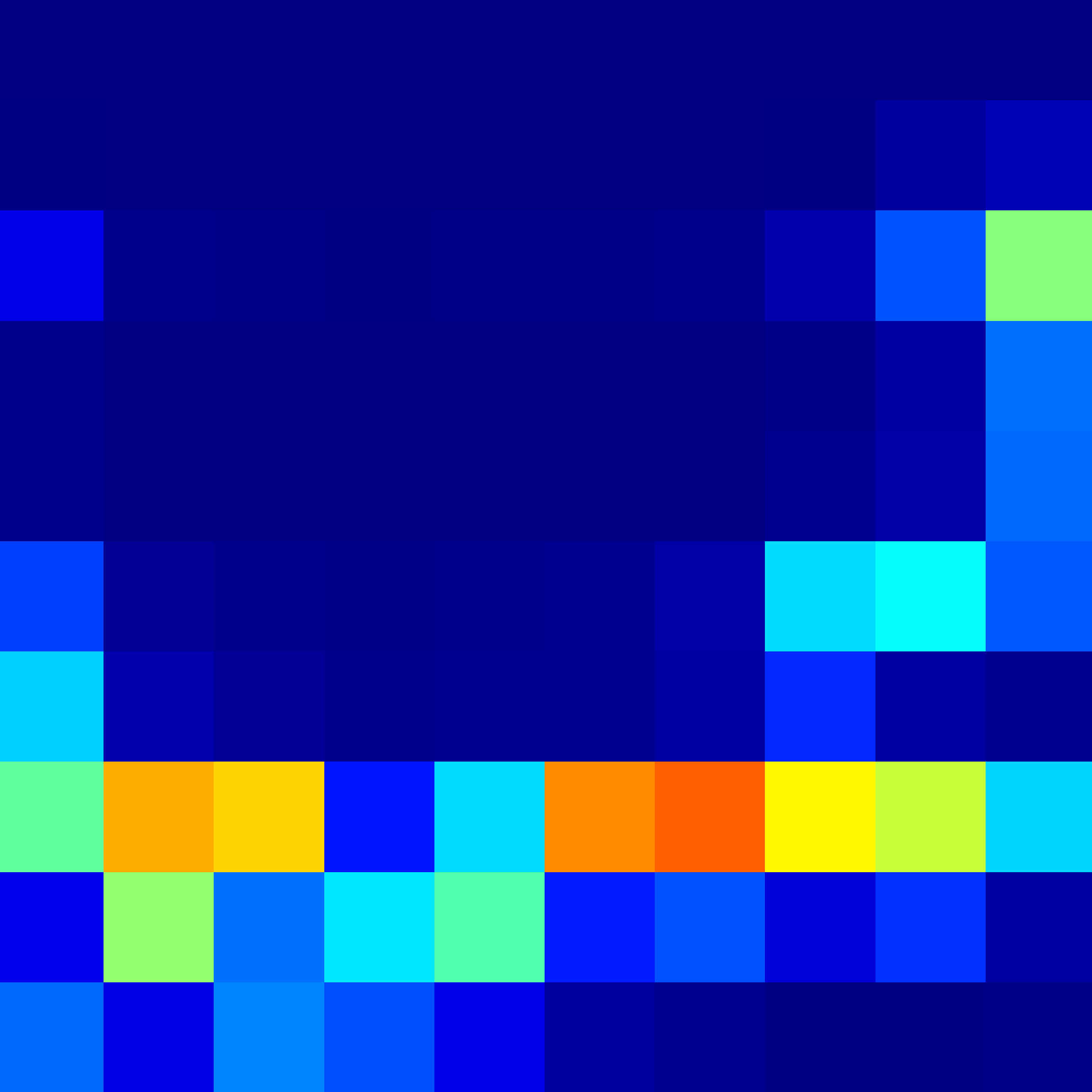}
    \caption{(\textbf{Left}) Tactile array with 100 taxels. (\textbf{Right}) Example of 10 x 10 tactile map from the array during manipulation. Red/orange regions indicate high contact forces and blue regions low contact forces.}
    \label{fig:tactile}
\end{figure}

\section{Experiments}

\subsection{Experimental Setup}

In our experiments, we employ a bimanual manipulation platform built on two LeRobot SO-101 manipulators~\cite{cadene2024lerobot}. The left is the standard SO-101 configuration, while the right arm is modified with a parallel gripper with tactile sensors on each finger. The left manipulator is actuated using 7.4\,V Feetech STS3215 servomotors, while the right arm employs the 12\,V version to provide increased torque required for precise grasping control.

As shown in Fig.~\ref{fig:robot}, the perception stack comprises three RGB streams. An Intel RealSense D435 camera provides the primary external view, and each manipulator is equipped with IMX335 5\,MP cameras mounted on the wrist. All cameras operate at a resolution of 640 × 480 pixels.

To enable contact-rich manipulation, we embed a parallel gripper with high-density tactile sensor arrays \cite{10.1007/978-3-030-22649-7_9}, shown in Fig.~\ref{fig:tactile}. An array of sensors is attached to each fingertip. A single electrode has 100 distinct taxels arranged in a 10 x 10 configuration. The range of force detection per point is 1--9\,N. Thus, we are able to receive tactile feedback with a resolution of 200 taxels in total at 120\,Hz. This high temporal and spatial resolution enables accurate monitoring of contact dynamics, including shifts in force distribution, incipient slip, and localized pressure variations, which are critical for safe and stable manipulation in contact-rich scenarios.

All computations are performed on an NVIDIA Jetson Orin NX 16\,GB edge computer.

\subsection{Experimental Tasks}

To evaluate the effectiveness of HapticVLA, we conducted experiments on three contact-rich manipulation tasks involving objects of different fragility and hardness: 
\begin{itemize}
    \item \textit{Jar pick-and-place}: "Pick a jar of marmalade and place it into the box", taking into account the high deformability of the plastic jar.
    \item \textit{Waffles pick-and-place}: "Pick waffles and place them into the box", which requires precise position and grasping force alignment due to the fragility of the waffles.
    \item \textit{Egg pick-and-place}: "Open the carton. Pick an egg and place it on the tray.", requiring even more precise grasp force and width regulation.
\end{itemize} 

For evaluation, we conducted 20 test trials for each task on each of the models. An episode is considered successful if the manipulated object remains completely intact and is placed at the designated target location. To account for task-specific characteristics, we additionally refine this criteria for each task individually:
\begin{itemize}
    \item \textit{Jar pick-and-place}: As the plastic jar is highly deformable, we deemed minor surface deformations insignificant during grasping, as long as the jar fully remained its initial shape after manipulation without any visible structural damage. 
    \item \textit{Waffles pick-and-place}: The waffles inside the package should remain fully intact without any cracks and crumbles.
    \item \textit{Egg pick-and-place}: The egg should remain completely intact after manipulation and be placed into the correct slot of the egg tray.
\end{itemize}

\subsection{Main Results}

To provide a fair evaluation of the proposed HapticVLA, we benchmark it against three representative policies: SmolVLA~\cite{shukor2025smolvla}, X-VLA~\cite{zheng2025x}, and VLA-0~\cite{goyal2025vla}. SmolVLA (0.45\,B), which serves as our foundation model, is a recent, lightweight VLA optimized for high-frequency edge deployment via asynchronous inference and a flow-matching action expert. X-VLA (0.9\,B) is chosen for its compact size and for being pretrained on large-scale, heterogeneous datasets, yielding state-of-the-art cross-embodiment adaptability. Including VLA-0 in baselines allows us to compare against a bigger model which outperformed SmolVLA specifically on the SO-101 manipulator~\cite{goyal2025vla}. For the experiments, we fine-tune all baselines on the same in-house dataset we used to fine-tune HapticVLA.

\begin{figure*}[t]
    \centering
    \includegraphics[width=\textwidth]{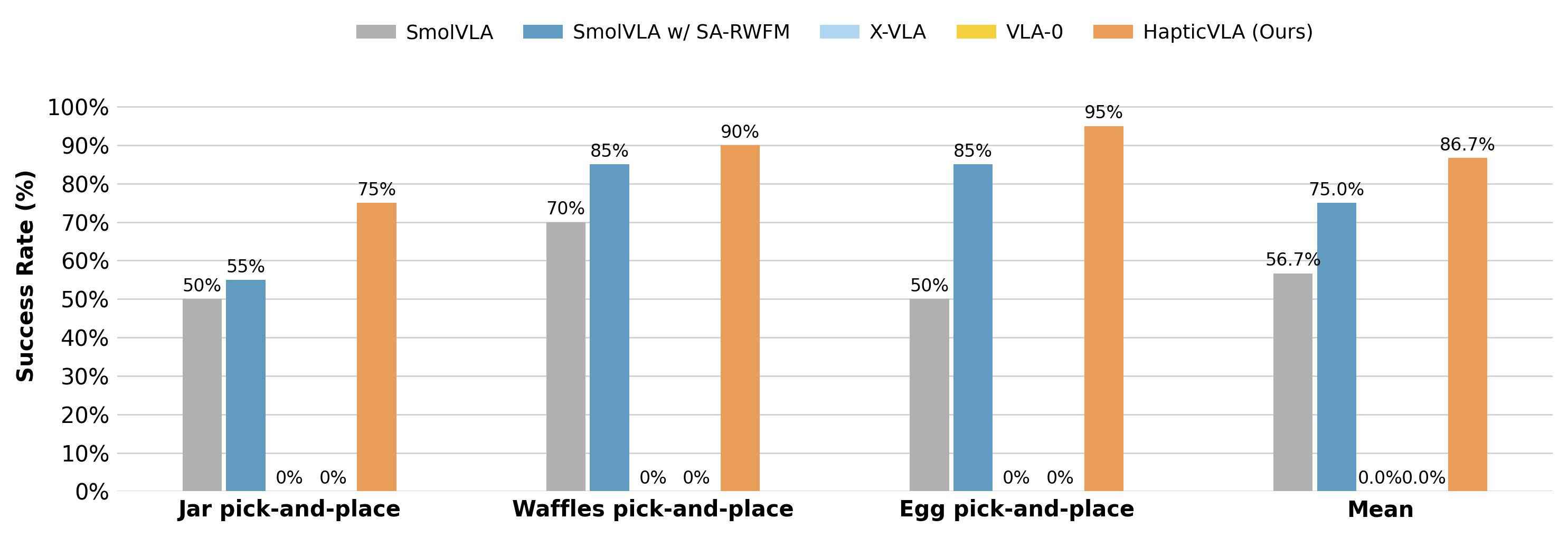}
    \caption{Success rates for three contact-rich manipulation tasks: Jar, Waffles, and Egg pick-and-place. Results are averaged over 20 evaluations per task. We compare HapticVLA (ours) with SmolVLA, SmolVLA with SA-RWFM action expert, X-VLA, and VLA-0. Our model consistently achieves higher performance, which highlights the efficacy of tactile distillation in combination with Safety-Aware Reward-Weighted Flow Matching for precise contact-rich manipulation.}
    \label{fig:experiments}
\end{figure*}

Fig.~\ref{fig:experiments} presents the success rate across three contact-rich manipulation tasks. HapticVLA significantly outperforms all baselines, achieving a mean success rate of 86.7\%. While the base SmolVLA struggles with the manipulation of extremely fragile objects like waffles and eggs, HapticVLA shows substantial performance gain: most notably, a 45\% absolute improvement over the base model in egg manipulation. SmolVLA integrated with our proposed SA-RWFM action expert yields a substantial performance leap, reaching a mean success rate of 75\% and achieving 85\% in two out of three tasks. Surprisingly, the zero-percent success rate of X-VLA and VLA-0 suggests that these models fail to generalize to the high-precision, contact-rich nature of our tasks. These results highlight the importance of haptic feedback integration into VLA manipulation and demonstrate the effectiveness of our approach in deploying tactile-aware manipulation policies in real-world settings.

\subsection{Ablation Study}

To verify the effectiveness of our proposed pipeline, we show the results of ablation experiments of our proposed Tactile Distillation. We evaluate four configurations on three contact-rich tasks, each with 20 independent test trials: 1) Without TD, with asynchronous SmolVLA inference, requiring tactile sensors at inference for contact-rich manipulation. 2) Without TD, tactile feedback is required at inference. 3) With TD, with asynchronous SmolVLA inference. 4) Our proposed HapticVLA with SA-RWFM and TD.

Table~\ref{tab:ablation} shows that TD substantially improves contact-rich manipulation performance and that combining TD with SA-RWFM yields the largest mean success rate across the three tasks. From the table, we could further find that asynchronous chunking of inference in some cases considerably lowers performance relative to the corresponding synchronous configurations. We attribute this degradation to temporal misalignment and increased effective latency between tactile observations and control actions.

\begin{table}[t]
\caption{Ablation Study of Tactile Distillation.}
\label{tab:ablation}
\centering
\begin{tabular*}{\columnwidth}{c@{\extracolsep{\fill}}cccc}
\toprule
Model & Jar & Waffles & Egg & Mean (\%) $\uparrow$ \\
\midrule
w/o TD, async & 16/20 & 18/20 & 15/20 & 81.7 \\
w/o TD & 11/20 & 17/20 & 17/20 & 75 \\
w/ TD, async & 14/20 & 19/20 & 15/20 & 80 \\
w/ TD & 15/20 & 18/20 & 19/20 & 86.7 \\
\bottomrule
\end{tabular*}
\end{table}

\section{Conclusion}

Considering the advancements in contact-rich VLA manipulation by integrating additional force and tactile modalities, we propose a novel pipeline, HapticVLA, tailored for contact-rich manipulation without requiring tactile sensors at inference. It enables tactile-aware manipulation via the proposed Safety-Aware Reward-Weighted Flow Matching (SA-RWFM) action expert, which accounts for precomputed tactile rewards in action generation. To enable inference of VLA with SA-RWFM without tactile sensors, HapticVLA employs Tactile Distillation (TD), which distills tactile representations from vision and state inputs. Experiments across diverse real-world tasks show that HapticVLA achieves a mean success rate of 86.7\% consistently outperforming baselines in manipulating fragile and soft objects without any haptic feedback.

\section*{Acknowledgments}
Research reported in this publication was financially supported by the RSF grant No. 24-41-02039.

The authors would like to express their sincere gratitude to Prof. Hiroyuki Kajimoto (The University of Electro-Communications) for developing and providing the tactile sensors employed in this work, and Dr. Miguel Altamirano Cabrera (Skolkovo Institute of Science and Technology) for his assistance in setting them up.

\bibliographystyle{IEEEtran}
\bibliography{references}

\end{document}